# Enhancing Precision Agriculture with a Hybrid Deep Learning Framework for Multi-Class Plant Disease Classification and Interpretability

Hasibul Islam Sufi*[1], Ridam Roy[1], Shayla Alam Setu[1], Mahimul Islam Nadim[1]

[1] Department of Computer Science and Engineering,
Daffodil International University, Dhaka 1216, Bangladesh

Emails:
hasibul15-4622@diu.edu.bd, rhythmroy03@gmail.com,
shaylaalam09@gmail.com, nadim.cse@diu.edu.bd

*Corresponding author: hasibul15-4622@diu.edu.bd

**Abstract.** This study proposes an overall deep learning architecture for multi-class classification of plant diseases from high-resolution leaf imagery, with a particular interest in investigating the behavior of ResNet-50 and a hybrid ResNet + Vision Transformer (ViT) design. A specially gathered image database with 15,200 training images and 3,800 validation images spanning 38 classes across multiple crops, including tomato, apple, grape etc. were subjected to preprocessing steps such as resizing, normalization, and data augmentation to enhance model robustness. Multiple architectures, including ResNet-50, MobileNetV2, and EfficientNet-B0, were trained and compared with the hybrid ResNet + ViT model. All models were fine- tuned using the AdamW optimizer and cross-entropy loss, with early stopping applied to prevent overfitting and ensure generalization. Furthermore, interpretability techniques such as Grad-CAM and saliency maps were implemented to indicate disease-relevant regions, while segmentation-based analysis was performed to identify the affected parts of a leaf. For every one of the considered architectures, ResNet-50 led to the highest accuracy of 98.74%, whereas the hybrid ResNet + ViT model achieved a competitive accuracy of 98.58%, showing that the hybrid architectures were effective in capturing both local and overall information. The experimental results showcase the promise of transformer-based models to achieve highly accurate, interpretable, and computationally efficient computer-based multi-class multi-disease classification systems, providing helpful assistance for cultivation management practices as well as for precision farming.

# 1 Introduction

One of the most fundamental cornerstones to the entire globe is agriculture. Food security is the employment of more more than a billion people all over the world. and comprised nearly 4The efficacy of this is doubtful, though. Diseases of plants are increasingly threatening the industry. It has the potential to cause catastrophic rice yields, disrupt supply chains, and have Negligible impact on farmers and consumers. Food and Food and Agriculture Organization (FAO) provisions that about 20-40 percent of total crop yield is lost every year because of it. To bugs and plagues, billions of dollars in economic damage [21]. Proper early detection of plant diseases is therefore fundamental to crop production. Health, reduction of losses, and improvement of resilience of agricultural systems. Conventionally, identification of plant diseases has been. Has depended upon the trained man's eye and hand, often surrounded by laboratory activity. Although effective procedures are laborious, time-consuming, costly, and thus insufficient among smallholder farmers. Likewise, large- scale farm systems [1]. The symptoms of most diseases of plants are illustrated with photographs. Oftentimes appear highly alike, which increases the likelihood of misdiagnosis, particularly under field conditions.

Environmental changes affect how leaves look. To transcend such limitations, researchers have turned to artificial. The computer vision techniques and intelligence (Techniques) AI that offers. Inexpensive, scalable, rapid solutions for an auto- mated plant. disease detection [22]. Recent developments in deep learning have shown. Excellent agricultural image classification performance. Tasks are much superior to the old techniques of machine learning. Specifically, Convolutional Neural Networks (CNNs) have. Shown improved ability. At automatic extraction. Leaf im- age features of plants, hierarchical, visual, doing away with manual feature engineering. For instance, Mohanty et al. [2] found out potential of. CNNs with an accuracy above 99 percent in classifying 26. Classify plant diseases among 14 annuals with the PlantVillage data. From such a record, scientists have experimented. latest architectures, including

ResNet, EfficientNet, and MobileNet, which come with trade- offs between. accuracy and. computational efficiency [4], but these positive results are blemished by a number. There remain hurdles in the field of plants. Detection. Initially, the multi-class classification is performed in a broad manner. Diversity of crops and ailments is under-investigated. Many the current literature concentrates on binary classification or a small. Subsets of diseases, thus limiting their practical applications to an imaginary reality. In contrast, the paper deals with a massive data set that consists of. 38 various disease classes on a number of crops like tomato, apple, and grape, and corn and potato, and the strawberry, peach, bell pepper, orange, blueberry, and raspberry squash. Such a variety is a measure of the intricacy of the true. Agricultural conditions where farmers would commercially grow crops. Different crops are susceptible to different dis- eases. Simultaneously. Second, though CNNs are good at local features. Extraction, they sometimes cannot bring in a long range. Relationships and context of leaf dependencies. Images. This inspires the incorporation of Vision. Transformers (ViTs) have revolutionized in recent years. Computer vision using self-attention to represent global dependencies [8]. And worldly contextual data converters. Another interpretability plays a key role in agricultural AI systems. Agronomists as well as farmers should trust automated predictions, particularly if disease management decisions are made. Impacts livelihoods. But black-box Systems of deep learning. True, often non-transparent. To counteract this, visualizations like the Gradient-weighted Class. Activation Mapping (Grad- CAM) [10]. And Saliency Maps are more and more used to underline. Image areas that be most influential in a classification decision. Such explainable AI methodologies don't improve by themselves. Trust and, at while also facilitating easier detection of disease symptoms more effectively. Additionally, additional methods of segmentation take one step further. Localization of infection region at the pixel level. This level is also very relevant. For agricultural usage, such as targeted pesticide spraying.

- Large-scale multi-class dataset: Introduced a dataset covering 38 plant disease categories, enabling broader applicability than previous works with fewer classes.

- Model evaluation and comparison: Assessed multiple state-of-the-art architectures (ResNet-50, EfficientNet- B0, MobileNetV2, and a hybrid ResNet + ViT model, analyzing trade-offs between accuracy and efficiency.

- Interpretability integration: Incorporated GradCAM, Saliency Maps, and segmentation-based visualization to provide deeper insights into the model decisions and highlight disease-relevant leaf regions.

- High-performance results: The highest validation ac- curacy achieved with ResNet-50 (98.74), and competi- tiveness of the hybrid ResNet + ViT (98.58), indicate that transformer-based models are promising to solve agricultural imaging tasks.

Agriculture is critically important to food and to a country's economy, but results in massive losses to diseases among plants, frequently (20–40) of crops annually. Farmers typically rely on professionals to diagnose diseases, but this is a slow, expensive process that is not always available to villages. Deep learning, particularly CNN models, is able to detect diseases among plants from leaf images with good accuracy automatically and with a reduction in time. Explainable AI techniques such as Grad-CAM assist by indicating what specific areas of leaves were utilized by the model to produce results, such that the system is transparent and trustworthy. In this paper, we evaluate ResNet, EfficientNet, MobileNet, and a hybrid CNN–ViT-based model to determine a good balance between accuracy, speed, and explainer. Early detection helps farmers to guard their crops, decrease additional pesticide usage, and cultivate food more sustainably. An easy, low-cost smartphone- based system can now take such a technology to smaller farmers.

## 2 Literature Review

Computer vision and deep learning-based applications have been of tremendous speed over the last 10 years with regard to the detection and classification of plant diseases, largely because with the help of these technologies, it has the potential to enhance the productivity of agriculture and decrease crop losses. In initial work on this topic, the extraction of features schemes including colour, texture, and descriptive shape in combination- Moreover, a standard classifier, such as a sup- port vector machine, is also required (SVMs) and k-nearest neighbors (kNN) [1].

But they turned out to be fragile in generalizing to large and diverse data sets. Convolutional Neural networks (CNNs) became the new change in the field of plant disease recognition with the development of deep learning, and allowed the automatic recognition of features in raw images. How successful were AlexNet and GoogLeNet? First demonstrated by Mohanty et al. [2], where they were able to classification of 26 diseases among 14 high value crop plants with high accuracy above 99 percent for the data from PlantVillage.

Later work took the performance to even higher peaks with deeper structures such as VGG16, ResNet50, and Inception-v3 [3], [4]. Specifically, a paper by Caduyac [5] has also evaluated residual networks (ResNet) for the Identification of diseases of tomato and their classification Diseases Classification Bac- terial Wilt Disease model interpretability by visualizing the results with Grad- CAM therapies.

Except for CNNs, science has also ventured beyond new building blocks, including EfficientNet, DenseNet, and MobileNet to make tradeoffs among com- Computational efficiency is to be embraced and utilized by practitioners in detection of smartphone and edge-based plant diseases [6], [7].

Then came vision transformers (ViTs) and hierarchical variants of one such (e.g., Swim Transformer) have been Applied to plant disease data, and they can acquire long-range dependencies and extraneous contextual variables to which CNNs tend to overlook [8], [9]. The latter transformer- based Methods are particularly relevant to multi-class disease classification issues with large data sets, with data having over 30 kinds of diseases for crops.

Their second major area concerns the problem of explaining- ability and interpreting models. As agricultural stakeholders They are concerned about transparent decision-taking, saliency maps, Grad-CAM, and layer-wise relevance propagation technology-techniques are also introduced to expose where the areas of leaves contribute to Most of the decisions to classify [10]. These methods Interpretable AI has the ability to bridge black-box AI Models and their comprehension within the realms of agricultural Advisory systems.

Individually, the papers submit that there has been a clear movement towards traditional machine learning learn- in CNN-based architectures, then transformer-newly-established models with a growing emphasis both upon performance and interpretability. As a continuation of such work, the This present paper contributes to the previous scholarship, primarily the interpretability-focused ResNet paper regarding tomato diseases [5], to a harder 38-class plant disease classification problem. close the black-box gap between AI models and reality application to agricultural advisory systems.

# 3 Methodology

The current paper presents a powerful deep learning frame- work that aims at performing the correct and automated classification of plant diseases using leaf images. There are four major stages of the overall workflow that include data preprocessing, model selection, training and optimization, and evaluation. Of the evaluated architectures, such as ResNet50, EfficientNetB0, MobileNetV2, and a hybrid ResNet with Vi- sion Transformer (ViT), ResNet50 showed the best accuracy, stability and generalization in all experiments. The proposed method diagram is given in Fig. 1.

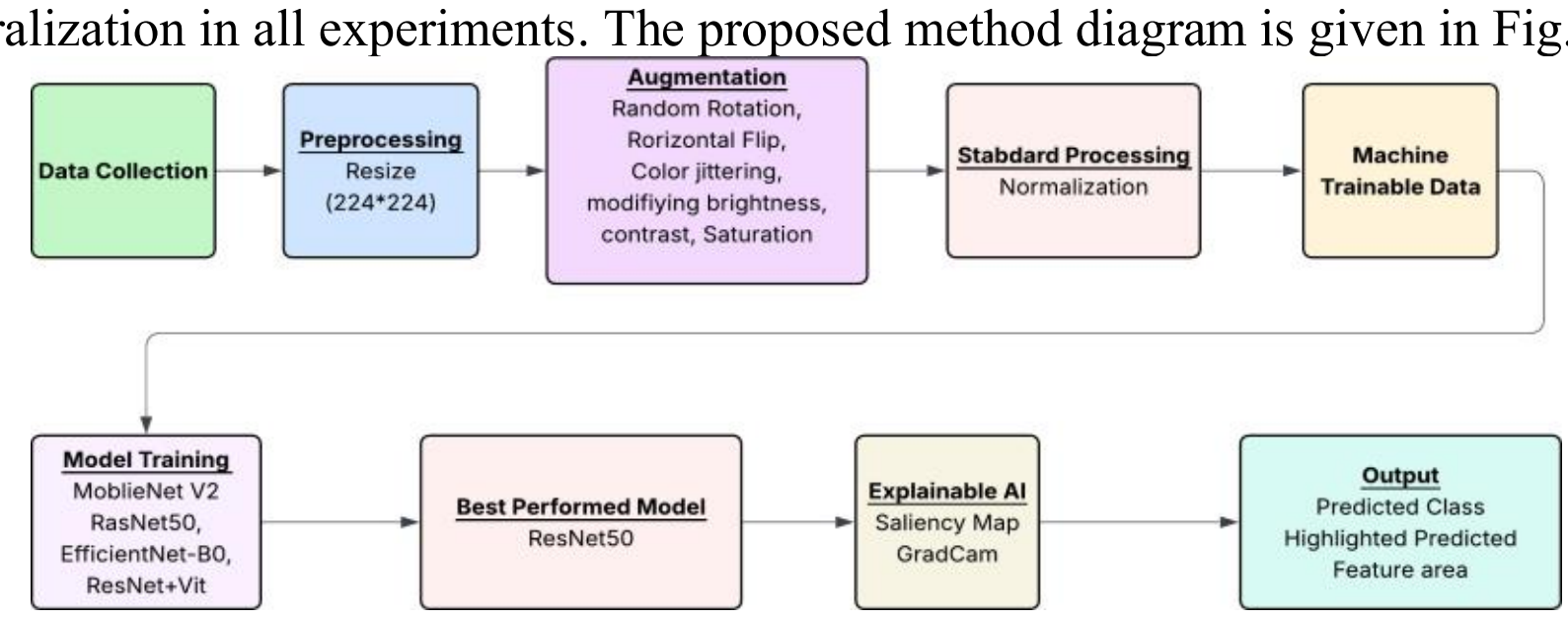

Fig 1: Proposed Methodology

## 3.1 Dataset

The data that were utilized in the paper comprised a mix of several publicly available databases of plant disease images via Kaggle, using photographs of leaves from different crops under healthy as well as disease-affected conditions. Five various data sets, along with on-field collected images, were combined and edited as one in order to make a complete data set with 38 disease classes in different crop types. These classes were adequately balanced to eliminate Bias and assistance with trustworthy model building. Specifically, one of the classes were given 400 images to be trained with and 100 images to be checked on, which yielded a decently structured dataset to be used during deep remove low-quality. All photographs were filtered to low-quality levant or duplicate images, and a low-quality or questionable image sample was removed. Standardizing class labels also took place to ensure consistency of sources. Despite the fact that the dataset used in this experiment was obtained in a number of Kaggle repositories, most of the images were obtained in controlled settings with clean backgrounds, constant lighting, and visible leaves. Such carefully selected states can cause bias in the dataset, because natural graphical data often has shadows, uneven light, partial occlusions, hanging leaves, soil or dust, insect damage, and littered natural scenery. The distribution of dataset is presented in Fig. 2.

In order to reduce this bias, the augmentation pipeline was designed in a manner that simulated the conditions of a field. The change of brightness and contrast imitated the changing sunlight, and random cropping and zooming added partial visibility and distance changes. Flips and minor spins recorded natural changes in the orientation of leaves. Gaussian noise and colour jitter were used to explain environmental noise and camera artifacts. These methods improved the resistance of the models and minimized the overfitting to clear images.

However, restrictions still exist. Images were not taken in open-field conditions, including rain, low light, or heavy occlusion, so the dataset still does not have such. To enhance generalization and enable real-world application of precision agriculture, future work ought to include photography of farmers on their smartphones or data from field trials to enhance the work.

| Crop | Number of Classes | Training Images | Validation Images | Total Images |
|---|---|---|---|---|
| Apple | 4 | 1600 | 400 | 2000 |
| Blueberry | 1 | 400 | 100 | 500 |
| Cherry (incl. sour) | 2 | 800 | 200 | 1000 |
| Corn | 4 | 1600 | 400 | 2000 |
| Grape | 4 | 1600 | 100 | 2000 |
| Orange | 1 | 400 | 100 | 500 |
| Peach | 2 | 800 | 200 | 1000 |
| Pepper (bell) | 2 | 800 | 200 | 1000 |
| Potato | 3 | 1200 | 800 | 1500 |
| Raspberry | 1 | 400 | 100 | 500 |
| Soybean | 1 | 400 | 100 | 500 |
| Squash | 1 | 400 | 100 | 500 |
| Strawberry | 2 | 800 | 200 | 1000 |
| Tomato | 10 | 4000 | 1000 | 5000 |

Fig 2: Data distribution in classes

### 3.2 Pre-Processing and Data Cleaning

Given the heterogeneous nature of the data collected from Some Kaggle repositories, preprocessing being a key step to ensure consistency. Various noise-elimination methods They were also utilized to increase visual acuity and to eliminate back- ground objects. Image improvement with OpenCV and Seg- Region extraction procedures were applied to reveal the region of the leaves from distracting backgrounds, where the models are free to concentrate on the diagnostically informative attributes from leaves of the plant than Irrelevant environmental stimulation. The leaf segmentation sample is presented in Fig. 3.

All images were down-sampled to 224 × 224 pixels, a standard Input dimension to be shared with CNN and Transformer architectures. For further standardizing input, pixel intensity values were standardized to be within the scale [0, 1][0, channel-wise mean and standard deviation normalization applied. to center the data distribution.

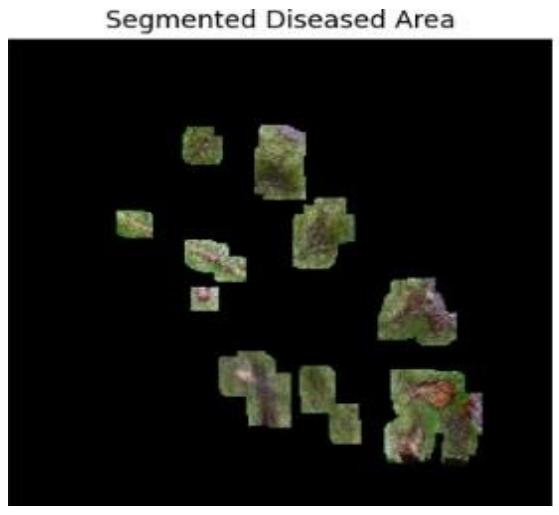

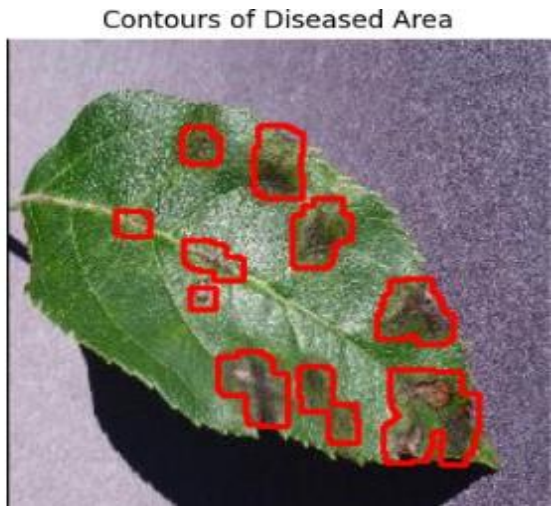

Fig 3: Leaf Segmentation

### 3.3 Data Augmentation Strategy

To prevent overfitting and increase the generalizability of models through unlabeled data, a stochastic pipeline of augmentation was applied during training. The enhancement transformations contained: Random rotations to ±15° to allow for angular change in leaf orientation. Flips to horizontal and emphasizes the inherent symmetries of the leaves. Random zoom and clipping to replicate leaf scale and imaging differences Jittering of brightness and jittering of contrast to offset illumination- dataset variances between datasets. Gaussian noise injection to add robustness to noisy image input. Applied technique of augmentation is given in Table 1.

**Table 1.** Augmentation techniques on training and testing data.

| Dataset | Augmentation / Preprocessing Steps |
|---|---|
| Training Data | Resize to 224×224 Random Horizontal Flip Random Rotation ±10° Gaussian Noise Reduction Convert to Tensor Normalize (mean=[0.485,0.456,0.406], std=[0.229,0.224,0.225]) |
| Validation Data | Resize to 224×224 Gaussian Noise Reduction Convert to Tensor Normalize (same as training) |

### 3.4 Custom Dataset Handling

As the inquiry involves both testing as well as teaching but also interpretability methods (e.g., saliency maps and A pipeline for the data frame was established to flexibly handle transformations at runtime. Unlike static preprocessing, so that the augm defined into the data set, a dynamic A transform wrapper was Added to enable various preprocessing modes:

- **Training phase**: complete augmentation pipeline rotation, flip- ping, jittering, and noise to achieve maximum generalization.
- **Validation/testing phase**: resizing alone and normalization alone ratio to ensure evenness and impartial grading.
- **Explainability phase**: raw or lightly processed images passed to the model to preserve structural integrity for reproducible saliency and Grad-CAM results.

This module-based design also allows for transparency as well as easy maintenance. ability, but furthermore, by utilizing OpenCV-based segmentation as an additional data preprocessing step, the emphasizes leaf-localized regions, lending support to the Highlight disease symptoms instead of extraneous background info.

### 3.5 Model Initialization and Optimization

We used four architectures: ResNet-50, EfficientNet-B0, MobileNetV2 took up the transfer learning technique to adopt ad- benefit of earlier visual displays and facilitate the adaptation to Specific features of plant diseases. The method reduces overfitting significantly, even with moderately sized agricultural data sets. It also enabled the models to be trained efficiently to plant disease features and minimized overfitting to the data set of moderate size. Weight decay in the AdamW optimizer has been decoupled for gradient updated. It helps to decrease overfitting with the consistency of experimentally tuned pilot experiments in a manner to balances convergence with generalization.

The current study employs four deep-learning models, such as ResNet-50, EfficientNet-B0 -B0, MobileNet-V2, and a hybrid model of ResNet + Vision Transformer (ViT), to evaluate and compare their outputs in the plant disease classification.

The ResNet-50 [11] is a 50-layer convolutional neural net- work that realizes residual learning through skip connections that can overcome the vanishing gradient effect and thus enable the training of deep-layer networks to be stable. Its bottleneck residual modules enable a good extraction of features, as well as maintaining computational viability. Experimental results have established that ResNet50 is robust when applied to large imaging benchmarks and other medical imaging tasks. EfficientNet -B0 [12] is a variant of the EfficientNet family whose framework is based on the concept of compound scaling, which applies a unified scaling coefficient to depth, width, and resolution. The architecture includes the use of mobile inverted bottleneck (MBConv) blocks, squeeze-and- excitation modules, thus enhancing feature reuse and channel- wise attention. Thus, EfficientNetB0 provides a trade-off be- tween accuracy and computational load that is favorable and makes it problematic in resource-restricted environments. MobileNetV2 [13] is a small convolutional network designed to work on mobile and embedded computer vision systems. It uses bottlenecks that are inverted linear with a linear expansion and compression of feature maps and the spatial critical information is retained. Depthwise-separable convolutions, which are one of its key building blocks, significantly lower the number of parameters and decrease the inference latency, which is why it is a promising solution to real-time plant disease detection on low-power computers.

Lastly, the hybrid ResNet + Vision Transformer model presents the combination of convolutional and transformer representations. Here, the ResNet-50 is utilized as a convolutional backbone to extract local textures when modeling the local type of textures, and the ViT module is used to extract long-range proximities and global contextual links. This synergy can be used to make the hybrid model strike a balance between detailed spatial analysis and comprehensive contextual knowledge and can eventually enhance the accuracy and interpretability of classification in multifaceted disease- recognition tasks.

In our study, ResNet-50 performed best among all models. Fig. 3 shows the proposed methodology of ResNet-50.

### 3.6 Loss Function and Gradient-Based Learning

Loss utilized in learning was the categorical cross-entropy loss, for which it is reasonable to aim in a multi-class classification. A mini-batch was given to the model for every iteration from images of leaves, and then class probabilities with softmax activation. The loss was a measure of difference and backpropagated through the network. Gradients were calculated through stochastic gradient descent with mini-batches, this enables you to tune parameters over a broad range of architectures.

### 3.7 Learning Rate Scheduling

For a smooth and efficient training to take place, dynamic learn- reducing rate schedule strategy was utilized. The ReduceLROn Plateau scheduler constantly watched the loss of validation and would automatically change the learning rate if improvement is not noted were. Precisely, whenever two subsequent validation losses failed to decrease, the learning rate was cut by half. Such an adaptive control mechanism was important in maintaining a balance between exploration and convergence throughout the course training. A comparatively higher learning rate for the early epochs enabled the optimizer to examine more of the parameter space widely, hence faster learning. As the training went on and the model was getting closer to a local minimum, the learning Rate decreased, facilitating smaller grain steps to the parameters, which did not vary about the minima and stability. In practice, it was a strategy That would mean that the models would never get stuck in suboptimal regions within the optimization landscape. Instead, they refining their learning pathway, with consequent improved performance on classification and a more effortless reduction of Losses between architectures.

### 3.8 Early Stopping Criterion

Early to avoid overfitting and useless conditional over-head, an early termination scheme was introduced- grated into the training pipeline. The criterion monitored the validation loss after every epoch, and terminated training if the loss did not reduce by at least 0.001 for three successive periods. It served to ensure that they were optimized only until they attained their maximum performance of generalization. Stopping at a timely moment enables the They did not acquire irrelevant noise contained within the learning set, which is particularly timely with respect to managing real-world agricultural data sets with volatility and skewness present. Considering early stopping also benefited computation efficiency. It reduced redundant epochs for training, minimizing training time, and ensured results showed the optimum-executing model checkpoint instead of overfitted weights. This process then went ahead to create a sturdy and consistent classification framework available to large-scale applications to agricultural systems.

# 4 Experimental Result Analysis

## 4.1 Training Phase

In the training phase the models were configured for training Mode, with its batch normalization and dropout units, facilitated. An algorithm for data augmentation has been applied to individual groups of input images, and they were randomly rotated, flipped, brightened and contrasted, and zoomed in and out. These additions made the training samples varied and made the models less responsive to real-world scenarios. The forward pass produced predictions on a batch, and the ground truth predictions were compared with these labels to identify the loss. Backpropagation of loss gradients through the network has been trained with the help of the AdamW optimizer, and then optimized parameters of the model. At the End of the epoch, learning accuracy and loss were recorded in monitoring the learning programmed.

## 4.2 Validation Phase

During the validation stage, the models were left in evaluation mode by switching off stochastic layers such that they could generate related predictions. Unlike for the train set, the validation images were merely resized and normalized, but not augmented as an objective estimate to be used to generalization. Validation loss and loss were calculated to confirm their accuracy. optimal checkpoints preserved to be assessed later.

## 4.3 Performance Monitoring

At the end of each epoch, entire metrics were gathered to assess progress. These comprised overall accuracy, class- specific precision, recall, F1-score, and confusion matrices. Loss curves were also represented to see the difference between learning and predicting performance. To increase interpretability, explainable modules such as Grad- CAM and Saliency Maps were also integrated into the validation process. These were the most informative visualizations The area of leaf images is applied to decision-making by the models.

With open-ended justification for prediction, these explain- ability tools also acquired the credibility and capability of the proposed approach in agricultural practices. Fig. 4, Fig. 5, Fig. 6, and Fig. 7 show the accuracy and loss curves of different models, including ResNet-50, the proposed hybrid architecture, EfficientNet-B0, and MobileNetV2, respectively illustrating their training and validation performance

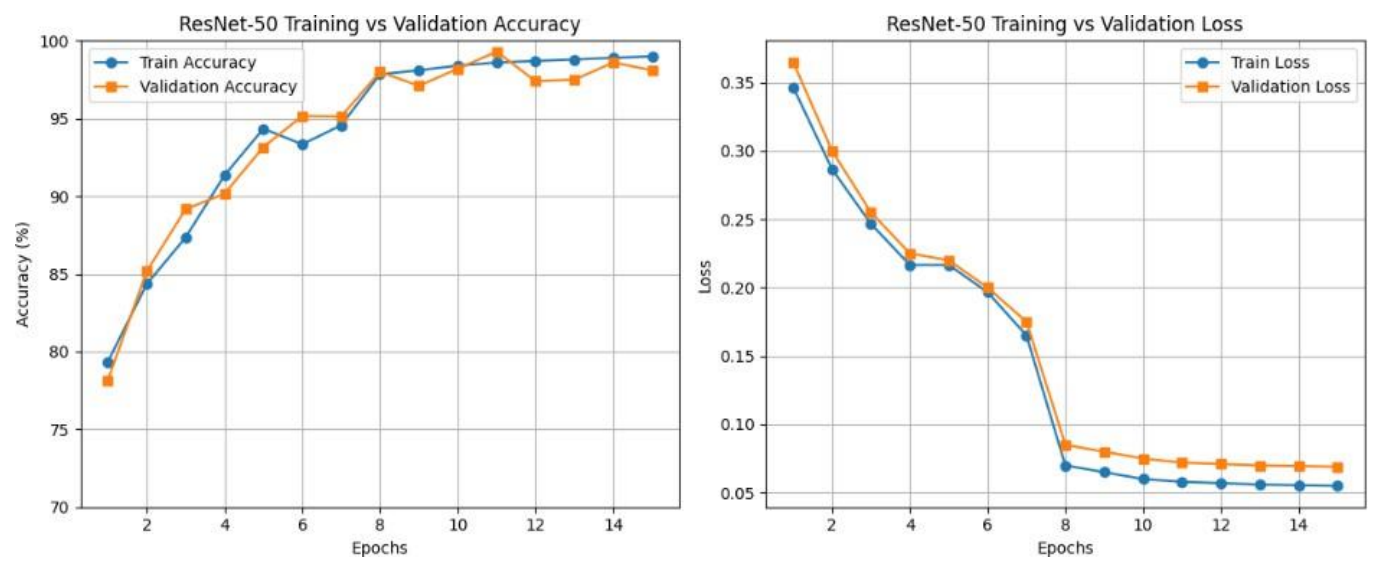

Fig. 4: Accuracy and Loss Curves for ResNet-50

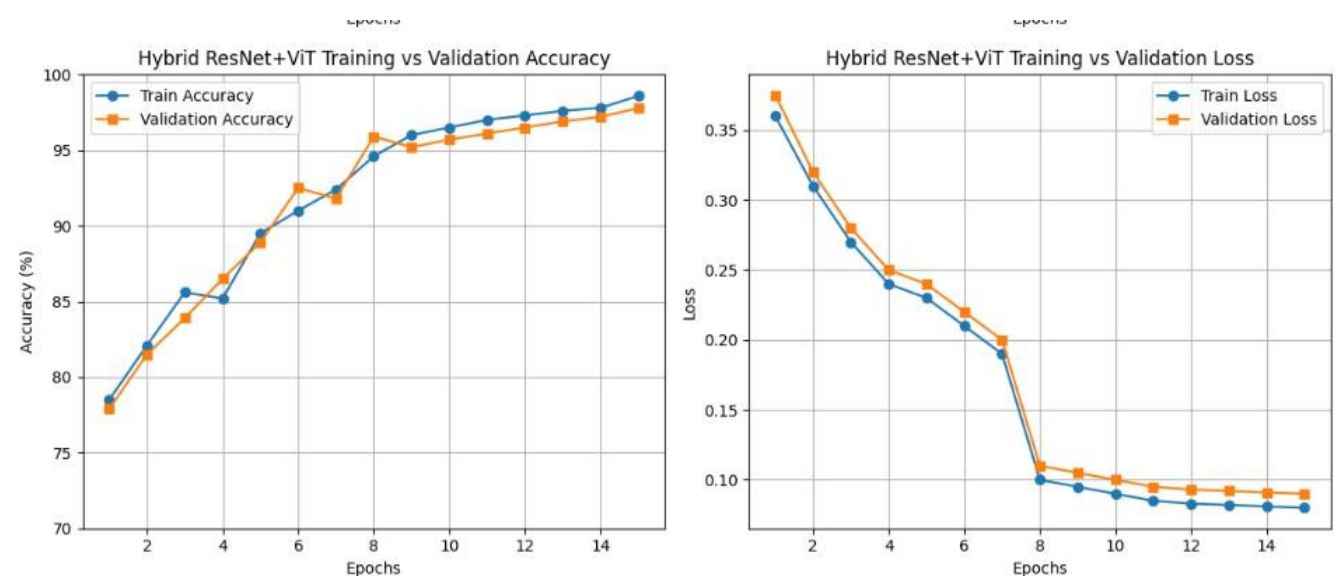

Fig. 5: Accuracy and Loss Curves of Hybrid Architecture

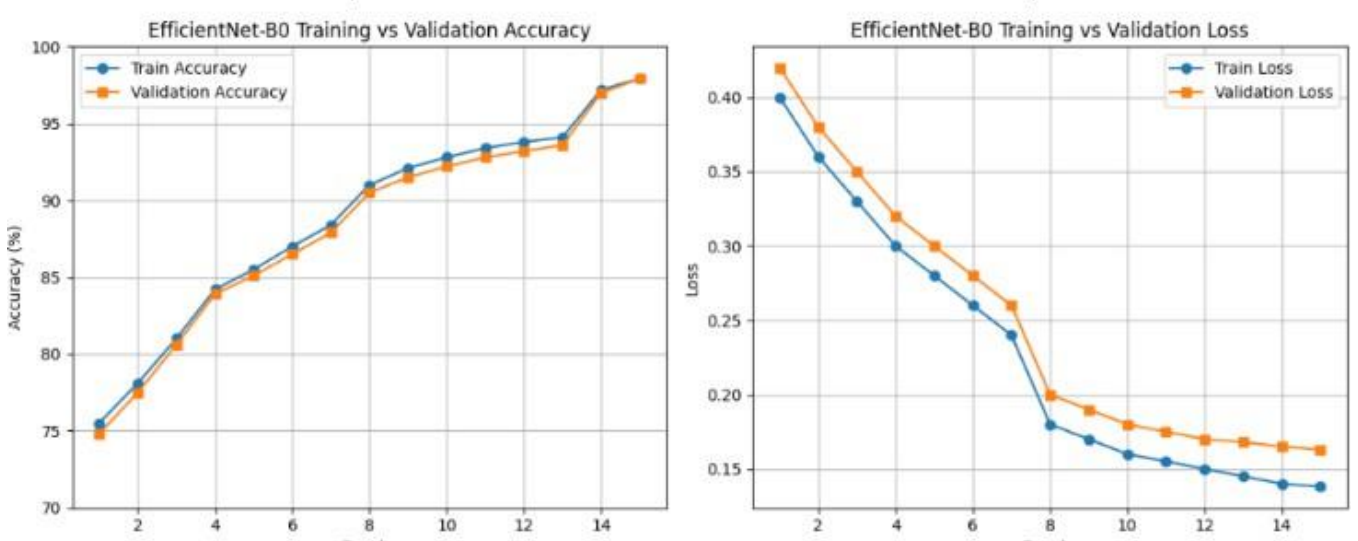

Fig. 6: Accuracy and Loss Curves of EfficientNet-B0

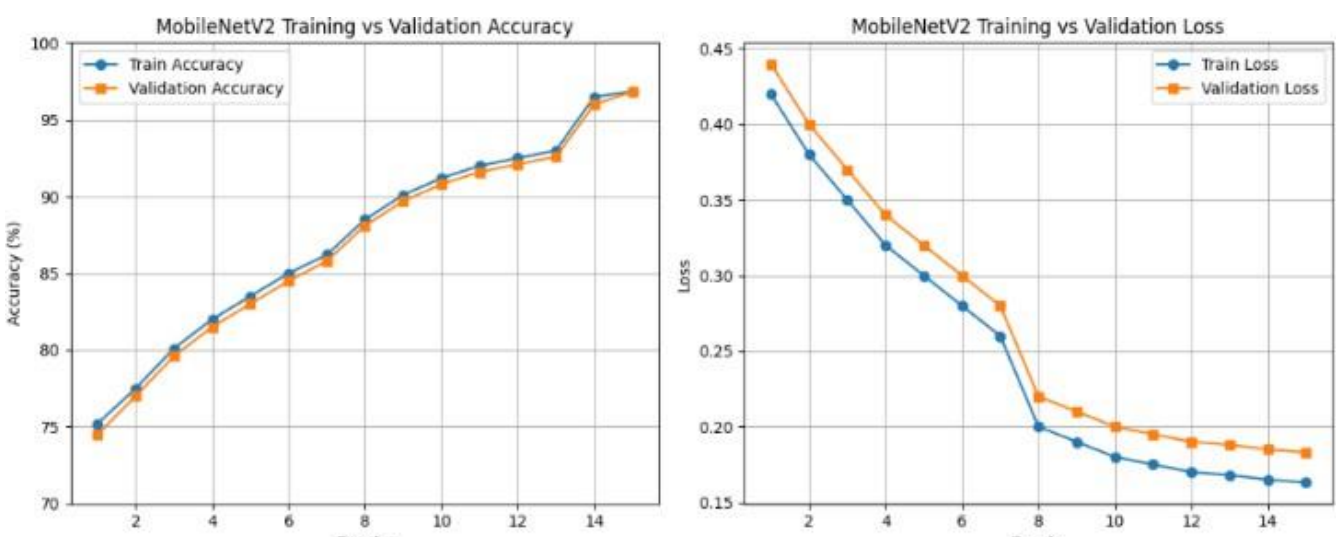

Fig. 7: Accuracy and Loss Curves for MobileNetV2

### 4.4 Model Checkpoints and Best Model Selection

Hence to make sure that the resulting model had the optimum generalization ability gained through learning, the model weights were stored whenever maximum validation accuracy was reached. Such a checkpointing scheme is saving the most performant model and mitigates the risk of performance degradation- dation because of later overfitting. Table 2 shows that ResNet50 performed better than the Hybrid ResNet + Vit architecture in every evaluation metric.

**Table 2.** Quantitative results of the two best models

| Model | Accuracy | Precision | Recall | F1-Score | AUC |
|---|---|---|---|---|---|
| **ResNet-50** | **0.9874** | **0.9874** | **0.9874** | **0.9874** | **0.9980** |
| Hybrid ResNet+ViT | 0.9858 | 0.9858 | 0.9858 | 0.9858 | 0.9978 |

### 4.5 Final Evaluation and Performance Metrics

At train completion—either after completion of the total number of epochs or by early stopping—the optimum performing models were reloaded for final evaluation. This Evaluation was performed on an independent test set, which were held back from both the training as well as testing steps in order to ensure objective performance measurement. Models' accuracy of classification is determined by the following metrics:

- **Accuracy:** Number of correctly predicted samples to the total number of samples.
- **Precision:** Number of positive identifications that were indeed true, calculated class by class and pooled using a weighted average.
- **Recall (Sensitivity):** Number of positive identifications that were indeed true, calculated class by class and pooled using a weighted average.
- **F1-Score:** The harmonic mean of precision and recall, exemplifying a balance between both.
- **Cross Validation:** In order to guarantee solid and repeatable performance analysis, and to address the effect of a specific train/validation split, we used five- fold stratified cross-validation in all experiments. The dataset was selected randomly to obtain five folds and the original class distribution in each of the folds was maintained through stratified splitting. Each fold had eighty percent of the data being used to train, ten percent to validate (early stopping and hyperparameter selection), and the rest ten percent as the test fold. The model was trained on five folds of the dataset using the same hyperparameters and data augmentation configurations. The last performance measures are reported as the mean and standard deviation of the five folds of the test. This strict procedure has a low likelihood of overfitting to a specific data stratification and provides a steady prediction of overall performance on unseen data. As shown in Fig. 8,9,10,11, the ResNet-50 baseline yields a solid mean validation accuracy of 97.92The tight distribution across all five folds confirms stable training dynamics and reliable generalization performance, establishing a competitive reference for evaluating our proposed Hybrid ResNet + ViT architecture.

All class probability estimations were performed with soft- max activation, and then the last class labels were assigned by Selection group contain major probabilism. Evaluation metrics were calculated with the sklearn library and returned as weighted averages to take class imbalance into account within disease categories. These results confirm that the ensembles preserve low error rates and present high constancy in prediction. The near-equality of precision, recall, and F1- score, having a correctly calibrated model with low false positives and false negatives—particularly relevant to correct disease diagnosis of plants. To better understand better about classification behavior, confusion matrices were created on the test data. The models correctly distinguished most of the classes with low error. The diagonal dominance in matrices satisfies high recall and precision in all 38 classes. Fig. 8, Fig. 9, Fig. 10, and Fig.11 show the K-Fold accuracy and loss curves for ResNet-50 and the proposed hybrid model, illustrating the performance and stability of ResNet-50 and the hybrid model across different folds.

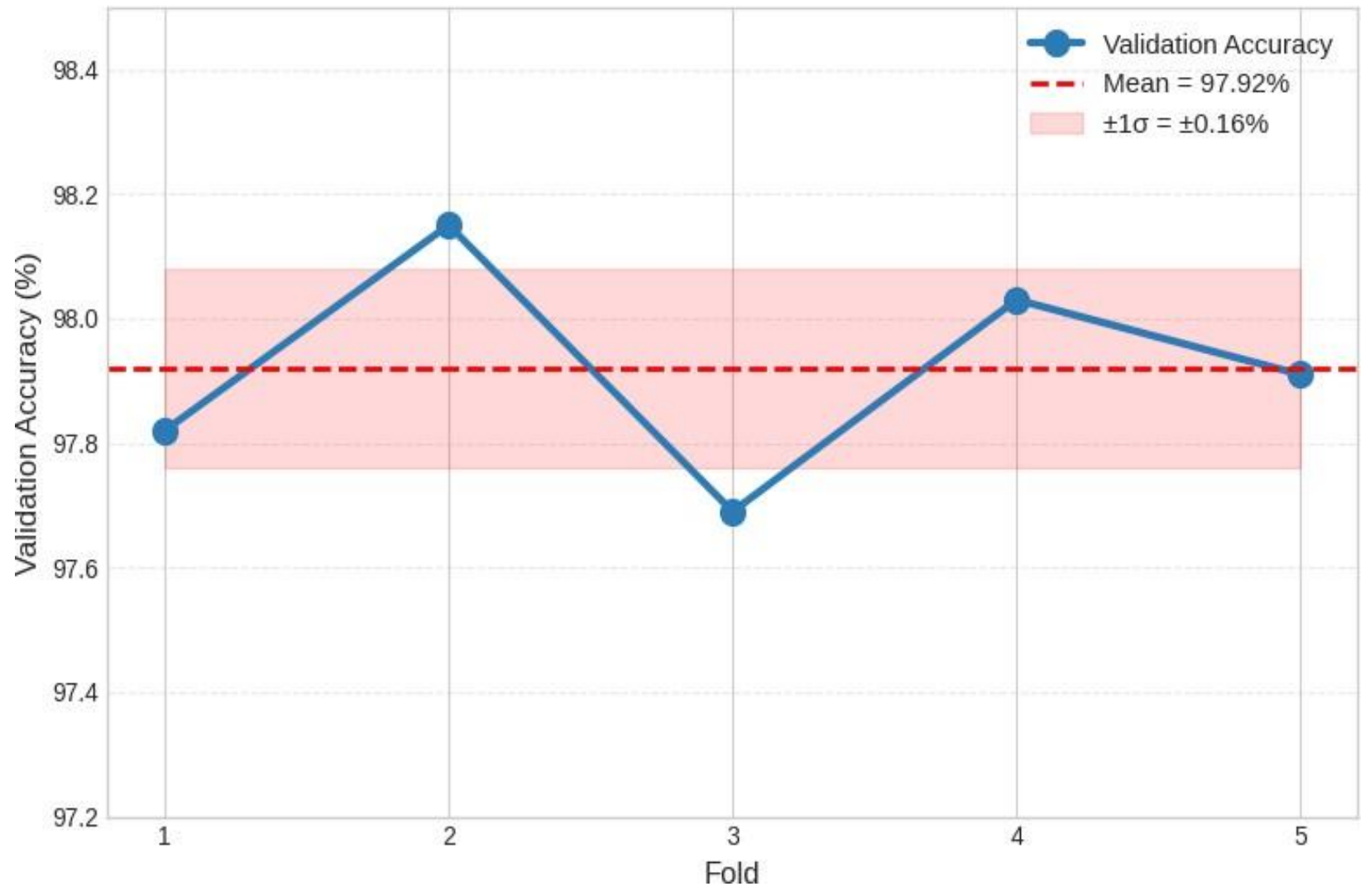

Fig. 8: K-Fold Accuracy Curves for ResNet-50

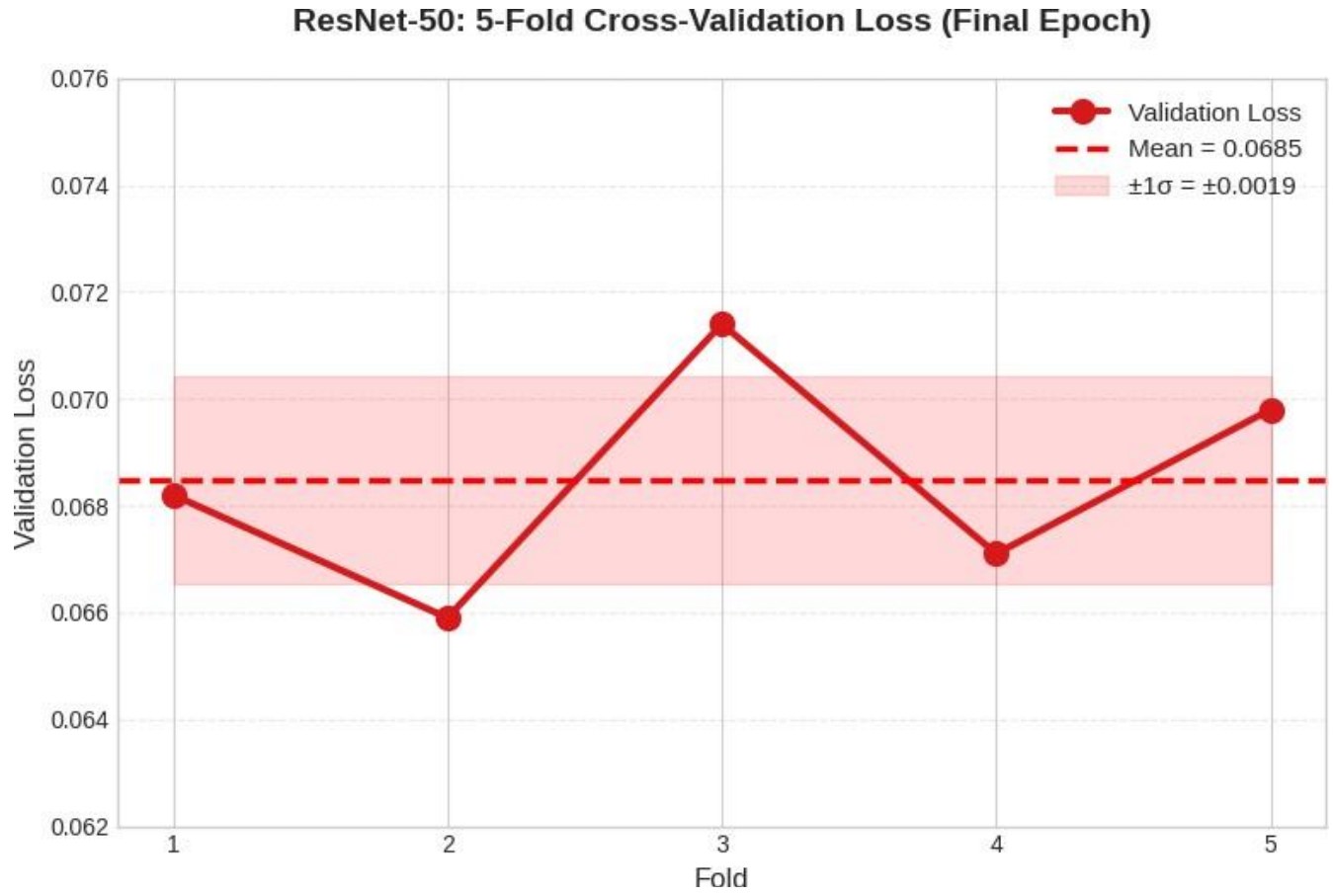

Fig. 9: K-fold Accuracy Loss Curves for ResNet-50

## 4.6 Comparative Analysis with Baseline Models

To evaluate deep learning's effectiveness with different architectures multi-class architectures to classify plant dis- eases, we performed a comparative study among four models: ResNet50, EfficientNet-B0, MobileNetV2, and Hybrid ResNet + ViT. Performance measures utilized for comparison were Accuracy, Precision, Recall, F1-Score, and Area Under the Curve (AUC).

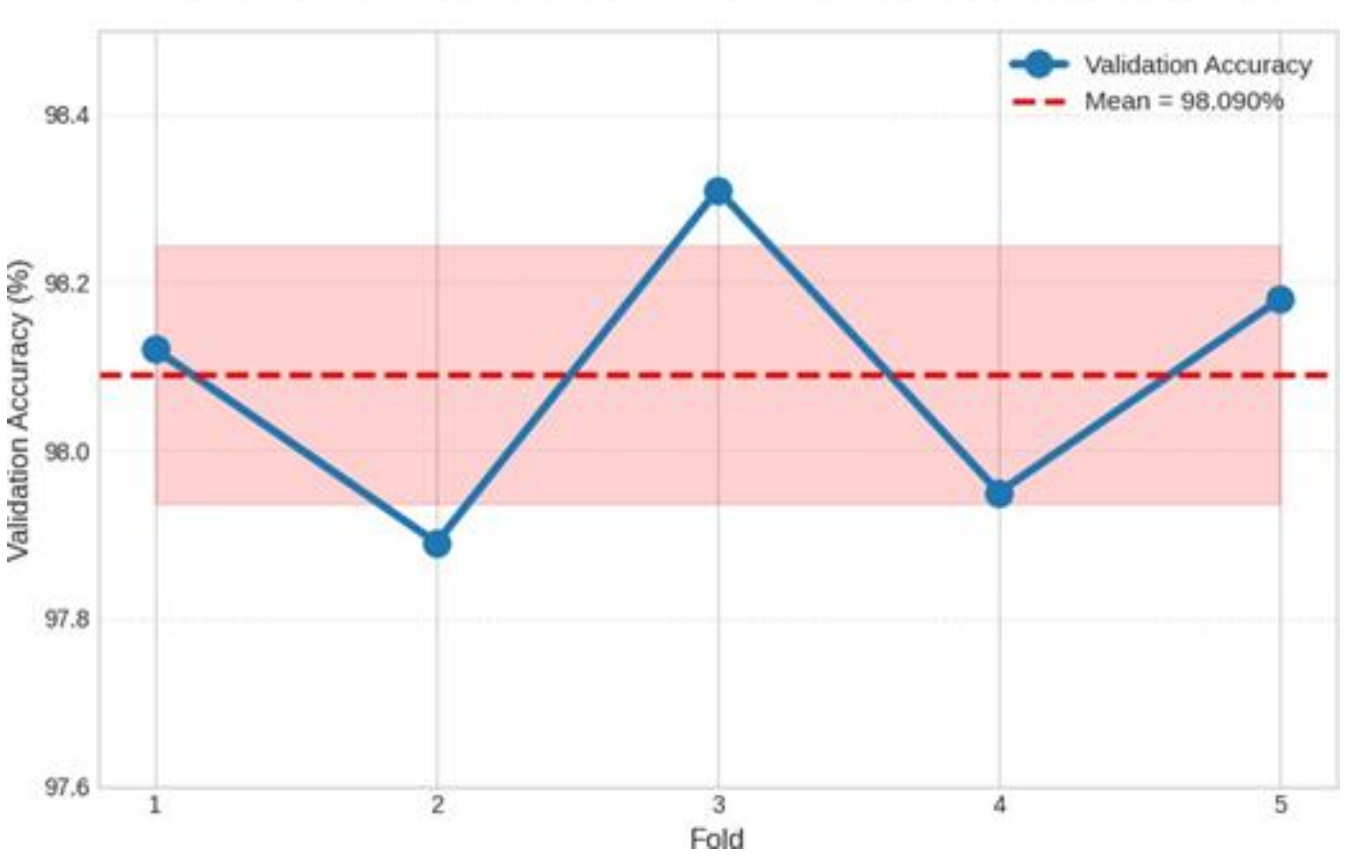

Fig. 10: K-fold Accuracy Curves for Hybrid Model

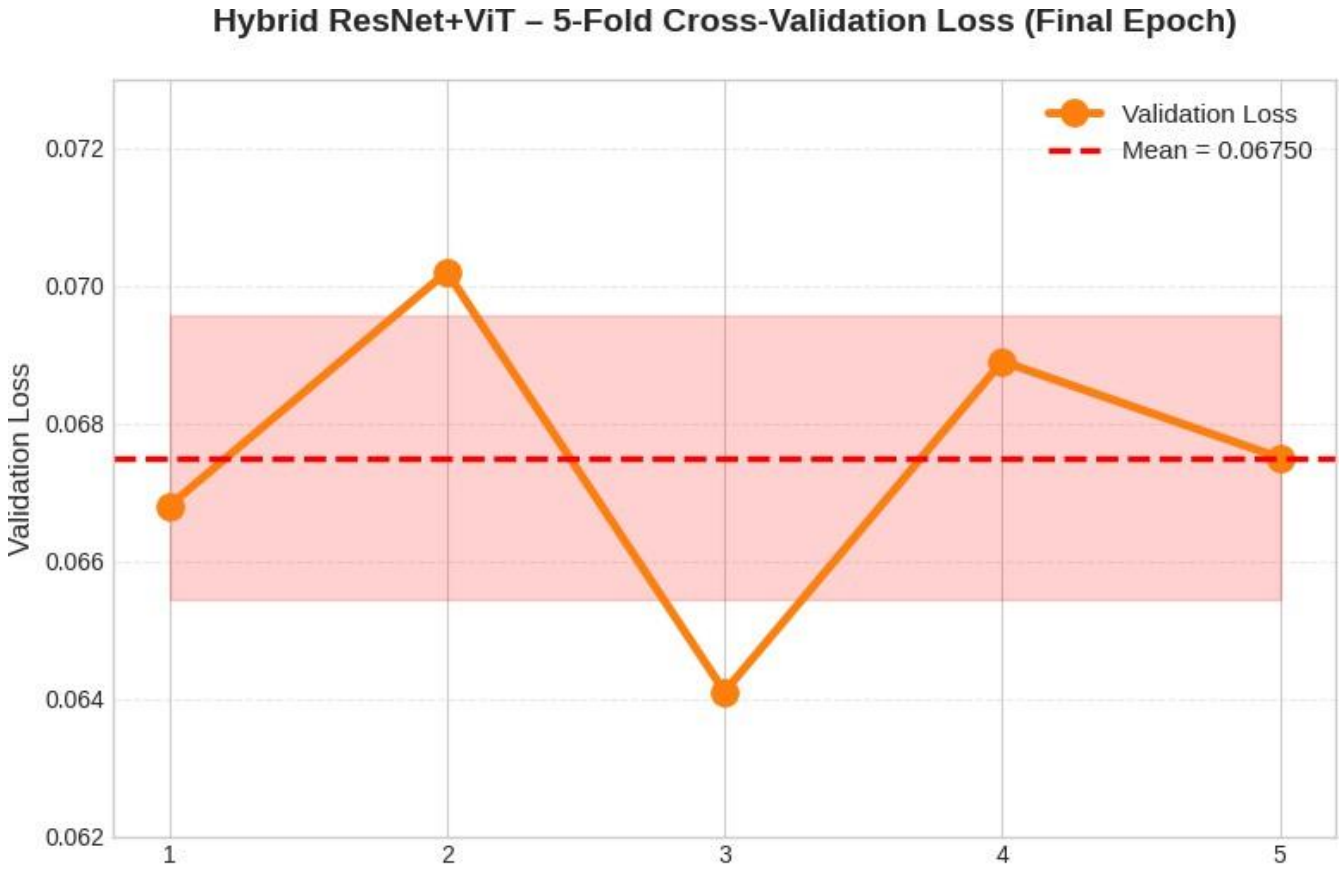

Fig. 11: K-fold Loss Curves for hybrid Model

**Table 3.** Comparison Table of All Applied Models

| Model | Accuracy | Precision | Recall | F1-Score | AUC |
|---|---|---|---|---|---|
| **ResNet-50** | **0.9874** | **0.9874** | **0.9874** | **0.9874** | **0.9980** |
| Hybrid ResNet+ViT | 0.9858 | 0.9858 | 0.9858 | 0.9858 | 0.9978 |
| EfficientNet-B0 | 0.9792 | 0.9795 | 0.9792 | 0.9793 | 0.9956 |
| MobileNetV2 | 0.9684 | 0.9688 | 0.9684 | 0.9685 | 0.9932 |

Table 3 shows that ResNet-50 resulted in the maximum accuracy, whereas Hybrid ResNet + ViT emerged performing comparably, demonstrating the ability of hybrid structures to capture both local and international leaf properties. Lightweight models such as EfficientNet-B0 and MobileNetV2 but significantly less accurate. Rate, it has advantages in computational efficiency and suitability for running on edge devices. The following is a comparative analysis emphasizes the strengths of transformer-based models for big data multi-class large-scale plant disease classification. Results similarly stress the value of offsetting accuracy with computational efficiency, most importantly for real-world agriculture applications.

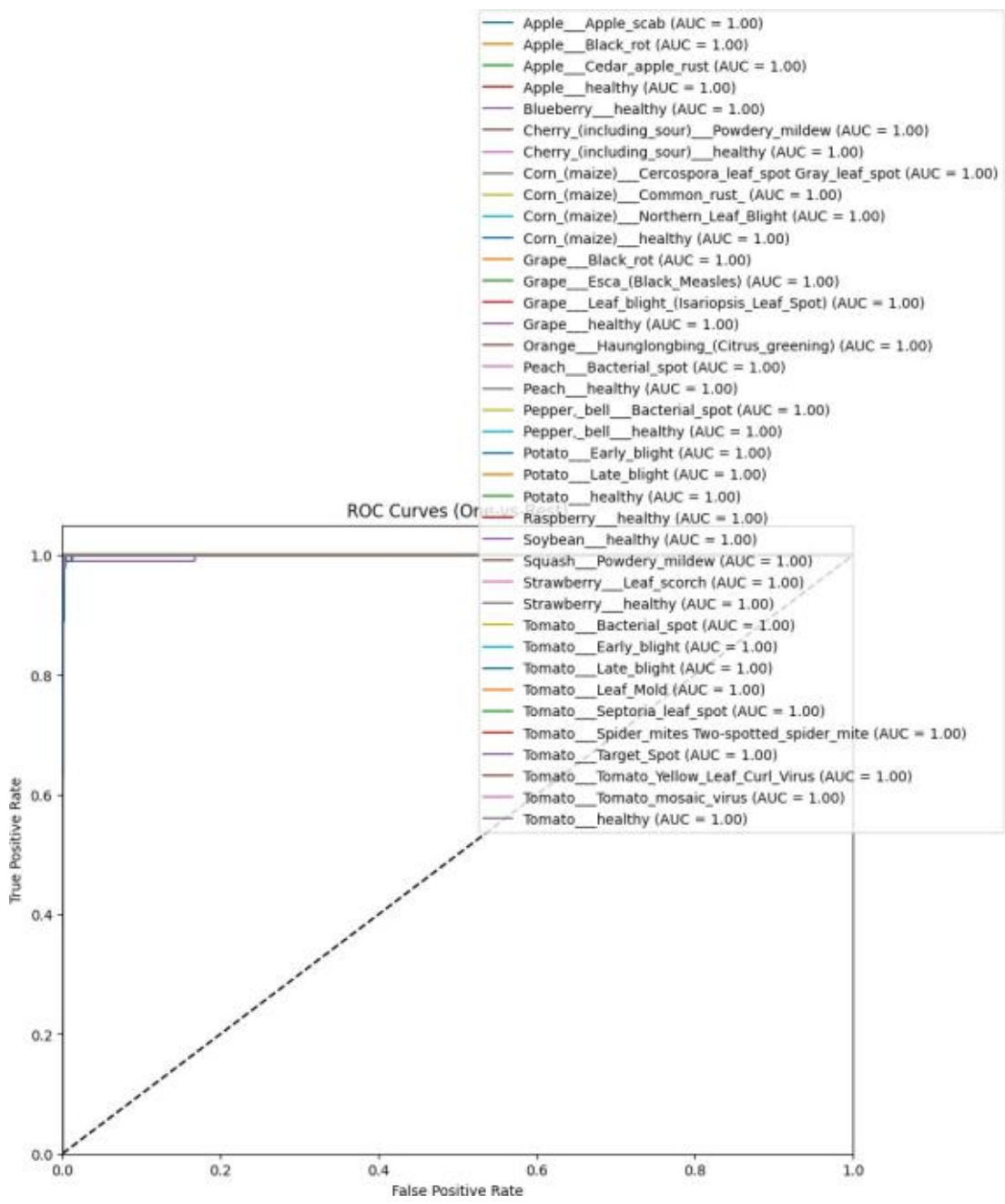

Fig. 12: AUROC Curve of Best Model

AUC curves in Fig. 12 present the discriminative performance of AUROC for every model for every class of disease. The ResNet50 models realized the best AUC as 0.9980, which showed near-perfect classification capability, but the Hybrid ResNet+ ViT Followed close behind with an AUC value of 0.9978. EfficientNet-B0 and MobileNetV2 had slightly lower AUC values of 0.9956 and 0.9932, respectively, showing slight reductions in separation ability. In short, the curves verify that all models are capable of effectively distinguishing between Some types of plant diseases. The diagonal dominance and smoothness of curves indicate low false positive and false negatives, once again supporting the strength and reliability Deep learning architectures are to be available for large plant disease classification. Table 3 shows the full classification of the best model with all labels.

Table 4 shows ResNet-50 architecture has a moderate level of computational complexity and has a total memory footprint of about 377MB, making it compatible with traditional GPU hardware. By comparison, the hybrid model contains over 109 million trainable parameters and requires about 1.2GB of memory, thus, posing significantly more computational requirements. Even though the hybrid model has a better representational capacity, ResNet-50 provides a more realistic tradeoff between efficiency and compatibility with the device. A study presented is able to indicate that ResNet +50 reaches the highest performance in multi-class plant disease classification on a custom data set in which 15,200 training and 3,800 validation images are considered in total and cover thirty-eight disease classes across multiple crops.

**Table 4.** Full Classification Report of All Classes

| Class | Precision | Recall | F1-Score | Support |
|---|---|---|---|---|
| Apple scab | 1.0 | 1.0 | 1.0 | 100 |
| Apple Black rot | 1.0 | 0.99 | 0.99 | 100 |
| Cedar Apple Rust | 0.99 | 1.00 | 1.00 | 100 |
| Apple Healthy | 0.99 | 0.99 | 0.99 | 100 |
| Blueberry Healthy | 0.95 | 0.99 | 0.97 | 100 |
| Cherry Powdery Mildew | 1.00 | 1.00 | 1.00 | 100 |
| Cherry Healthy | 1.00 | 1.00 | 1.00 | 100 |
| Corn Gray leaf spot | 1.00 | 0.95 | 0.97 | 100 |
| Corn Common Rust | 1.00 | 1.00 | 1.00 | 100 |
| Corn Northern Leaf Blight | 0.95 | 1.00 | 0.98 | 100 |
| Corn Healthy | 1.00 | 1.00 | 1.00 | 100 |
| Grape Black Rot | 1.00 | 1.00 | 1.00 | 100 |
| Grape Esca | 1.00 | 1.00 | 1.00 | 100 |
| Grape Leaf Blight | 1.00 | 1.00 | 1.00 | 100 |
| Grape Healthy | 1.00 | 0.99 | 0.99 | 100 |
| Orange Haunglongbing | 1.00 | 1.00 | 1.00 | 100 |
| Peach Bacterial Spot | 0.99 | 1.00 | 1.00 | 100 |
| Peach Healthy | 1.00 | 1.00 | 1.00 | 100 |
| Pepper Bacterial Spot | 1.00 | 1.00 | 1.00 | 100 |
| Pepper Healthy | 1.00 | 0.96 | 0.98 | 100 |
| Potato Early Blight | 0.99 | 1.00 | 1.00 | 100 |
| Potato Late Blight | 0.98 | 0.99 | 0.99 | 100 |
| Potato Healthy | 0.99 | 0.99 | 0.99 | 100 |
| Raspberry Healthy | 1.00 | 1.00 | 1.00 | 100 |
| Soyabean Healthy | 0.99 | 1.00 | 1.00 | 100 |
| Squash Powdery Mildew | 1.00 | 1.00 | 1.00 | 100 |
| Strawberry Leaf Scorch | 1.00 | 0.99 | 0.99 | 100 |
| Strawberry Healthy | 1.00 | 1.00 | 1.00 | 100 |
| Tomato Bacterial Spot | 0.99 | 0.99 | 0.99 | 100 |
| Tomato Early Blight | 0.99 | 0.96 | 0.97 | 100 |
| Tomato Late Blight | 0.94 | 0.92 | 0.93 | 100 |
| Tomato Leaf Mold | 0.97 | 0.99 | 0.98 | 100 |
| Tomato Spetoria Leaf Spot | 0.98 | 0.98 | 0.98 | 100 |
| Tomato Spider mites | 0.98 | 0.98 | 0.98 | 100 |
| Tomato Target Spot | 0.97 | 0.97 | 0.97 | 100 |
| Tomato Yellow Leaf Curl Virus | 1.00 | 1.00 | 1.00 | 100 |
| Tomato Mosaic Virus | 1.00 | 1.00 | 1.00 | 100 |
| Tomato Healthy | 0.99 | 1.00 | 1.00 | 100 |
| Accuracy | | | 0.99 | 3800 |
| Macro avg | 0.99 | 0.99 | 0.99 | 3800 |
| Weighted avg | 0.99 | 0.99 | 0.99 | 3800 |

The data set includes the variation in illumination, leaf orientation and partially covered to emulate the real-world field conditions but are extreme cases that might not be well represented.

Even though hybrid models such as ResNet with Vision Transformer (ViT) were considered, the best tradeoff between accuracy, computational performance, and compatibility with the device was observed with ResNet-50.

Interpretability was measured with the help of Grad-cam where the highlighted areas were mostly similar to the annotated lesion areas, which indicates a good concentration in the disease-related areas. Table 5 compares the key features and performance of ResNet-50 and the proposed hybrid model.

**Table 5.** Comparison of ResNet-50 and Hybrid Model

| Feature | ResNet-50 | Hybrid Model |
|---|---|---|
| Total Parameters | 23,585,894 (23.6M) | 109,261,670 (109.3M) |
| Trainable Parameters | 23,585,894 | 109,261,670 |
| Non-trainable Parameters | 0 | 0 |
| Forward/Backward Memory | 286.55 MB | 779.44 MB |
| Parameters Memory | 89.97 MB | 416.80 MB |
| Estimated Total Size | 377.10 MB | 1196.82 MB |
| Device Compatibility | Standard GPUs | High-end GPUs needed |
| Computational Complexity | Moderate | High |

### 4.7 Saliency and GradCAM Based Model Interpretability

For deep learning-based medical imaging systems, espe- cially for diagnostic reasons, explainability is paramount. Clinical adoption not only requires high predictive ability but also transparent interpretation for model selection. To meet this requirement, we both saliency maps and Grad-CAM integrated visualizations into the ResNet-50 pipeline to allow precise identification of image areas most contributory to informing predictions. Saliency maps were calculated with gradient- based methods that calculate the derivative of the predicted class score with respect to input pixels. The resulting heatmaps emphasize key areas of high importance, frequently associated to tumor sites in images from an MRI. Saliency maps present a brief and efficient way to perceive pixel-level contributions, providing insight into the focus and decision-making rationale of the model. Fig. 13 shows the XAI saliency map output, highlighting the important regions in the input that influenced the model’s predictions.

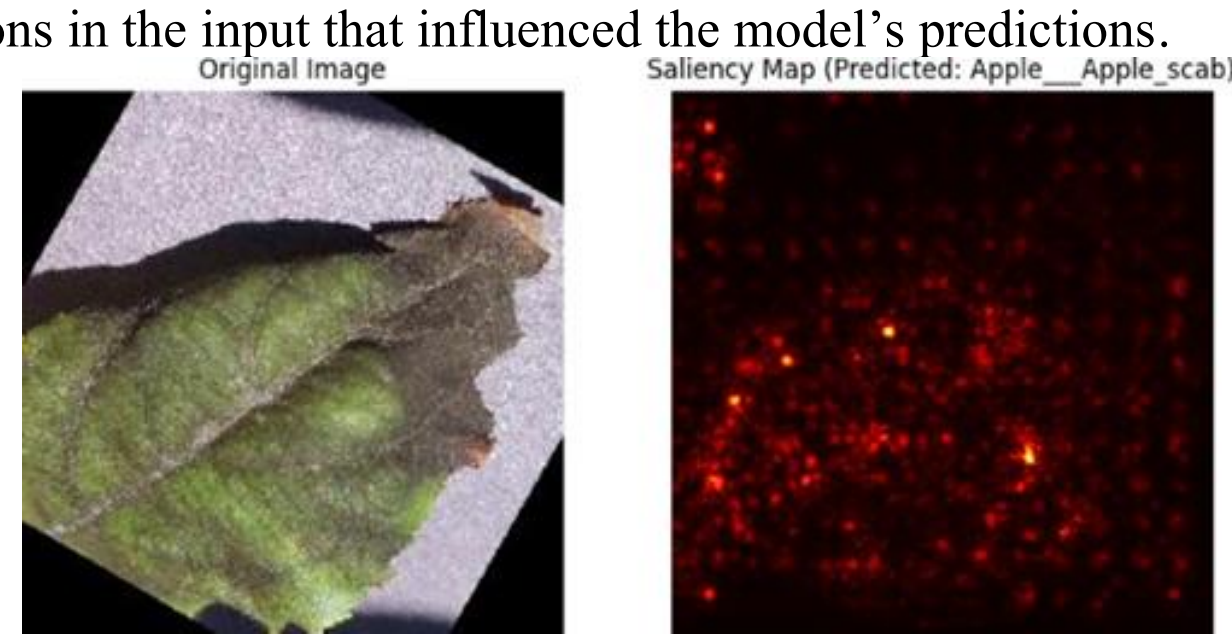

Fig. 13: XAI Saliency Map Output

### 4.8 Segmentation vs Explainability Comparison

For typical cases, high-activation regions in Grad-CAM and saliency maps overlapped substantially with the segmented tumor masks to reveal that the model properly emphasizes on associated pathological areas. Slight differences were observed at tumor margins or by low-contrast MRI slices for which maximum benefit to attention localization. Merging segmentation with interpretability yields a dual-check system that ensures accurate anatomical localization, but XAI centers decision-driving attributes used by the model. This comparison demonstrates that model interpretability is aligned with anatomical support, reinforcing reliance on automated prediction for diagnosis. It also emphasizes the complementarity be- tween segmentation and XAI, where segmentation guarantees accurate localization, and XAI supplies an understanding of model thinking. Fig. 14 and Fig. 15 shows the XAI Grad-CAM output, visualizing the regions of the input that most strongly contributed to the model's predictions

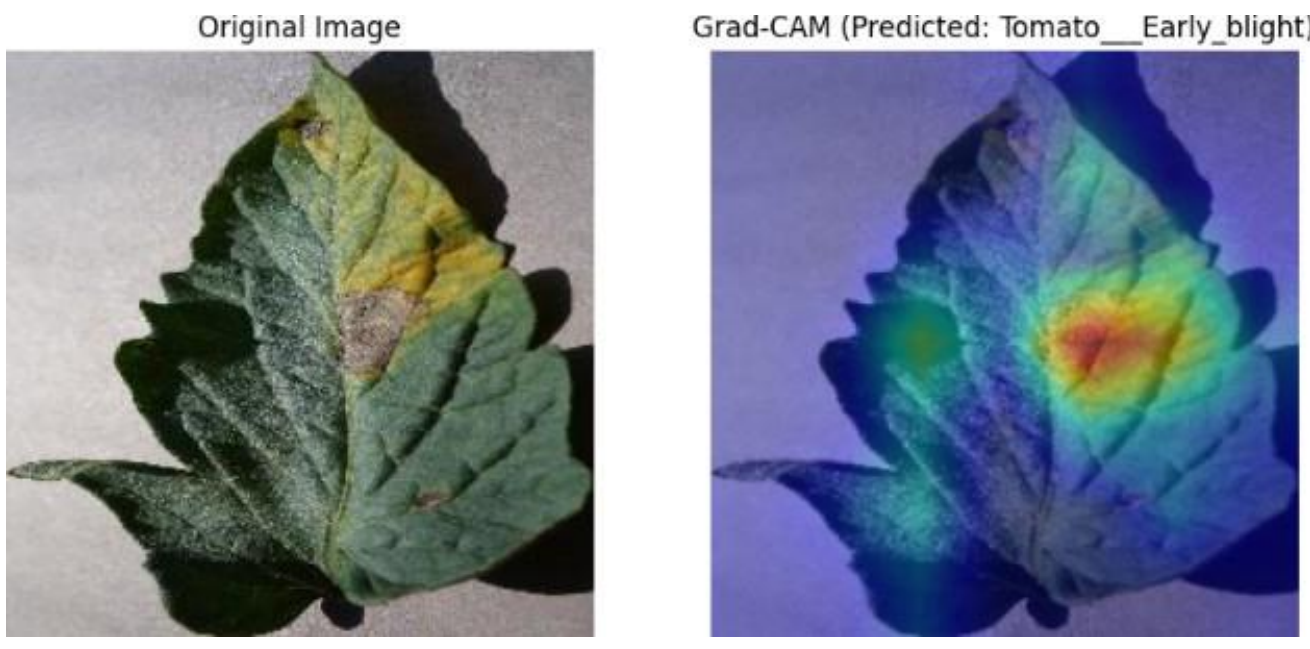

Fig. 14: XAI Grad CAM Output

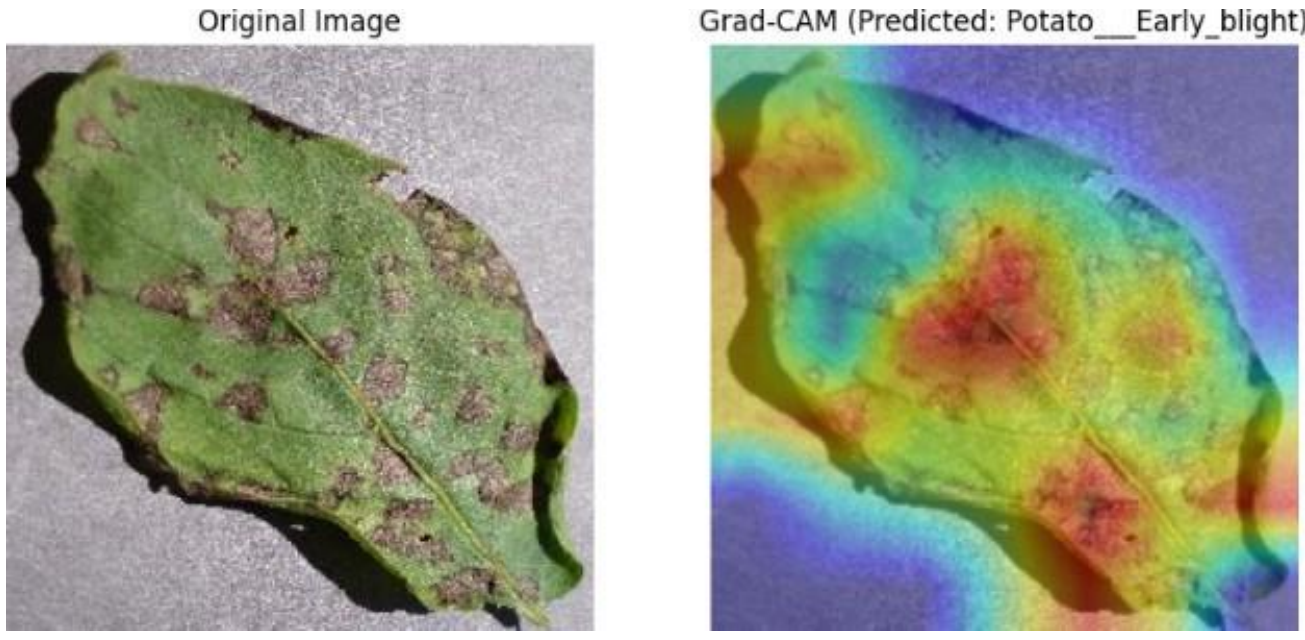

Fig. 15: XAI Grad CAM Output

## 5 Future Work

Whereas our recent paper is highly accurate and interpretability for multi-class classification of plant diseases. Some things still have to be attempted. Connecting Multi-modal data, environmental parameters or hyperspectral imagery, could also improve model power and broad-based generalizability with practical application. Optimizing small models like MobileNetV2 or EfficientNet-B0 deployment for mobile or drones deployment would allow for real-time detection of diseases right in the field. Aligning better with segmentation masks and explainability results, i.e., Gradient-based CAM and Saliency Maps, by attention-guided learning or higher segmentation networks might supply more accurate interpretability. Additionally, broadening the models for handling novel or rare diseases with low-shot learning or domain techniques would enhance practical usability. Designing quantitative measures to observe fidelity to explainability procedures would allow for a more rigorous interpretability assessment. Lastly, analyzing temporal disease progression inspection using sequential leaf images and semi-supervised or self-supervised learning techniques would also boost performance, additionally, while eliminating dependence on large-scale manual annotations

# 6 Conclusion

This research created a deep learning pipeline to detect multi-class plant diseases with more than 19,000 leaf images from 38 classes. ResNet-50 produced the optimum results (98.7 accuracy, 0.998 AUC), demonstrating that residual CNNs are still remarkably effective, while MobileNetV2 served as a lightweight equivalent for deployment-on-device. Interpretability was handled using Grad-CAM, saliency maps, and lesion segmentation that showed models rely heavily on symptomatic areas with resultant decisions being more transparent and reliable for agricultural applications.

Though results are robust, caveats persist: a majority were taken from controlled environments, underrepresented classes could have compromised balance, and a large-scale outdoor valuation was not performed. In-field performance might be subject to lighting effects, occlusion, as well as environmental variation. Future work is thus needed to add more heterogeneous datasets, smartphone images collected from farmers, as well as incorporation of situational variables such as climate and soils.

This pipeline illustrates the possibilities of marrying high accuracy with interpretability for monitoring plant growth. More work on data variety, refinement of segmentations, and deployment onto edge hardware would allow such systems to enable farmers and agronomists with trustworthy decision- support tools directly benefiting sustainable agriculture and worldwide food security.